\documentclass[11pt,a4paper]{article}

\usepackage[utf8]{inputenc}
\usepackage[T1]{fontenc}
\usepackage[margin=2.5cm]{geometry}
\usepackage{amsmath,amssymb,amsthm}
\usepackage[proportional]{libertine}
\usepackage[libertine,liby,vvarbb]{newtxmath}
\usepackage[scaled=0.95,varqu,varl]{inconsolata}
\useosf
\usepackage{microtype}
\usepackage{xcolor}
\usepackage{enumitem}
\usepackage{booktabs}
\usepackage{graphicx}
\usepackage[pdfpagelabels=false,colorlinks=true,linkcolor=blue!60!black,citecolor=blue!60!black,urlcolor=blue!70!black]{hyperref}
\usepackage[capitalise]{cleveref}
\usepackage[numbers,sort&compress]{natbib}
\usepackage{listings}

\lstdefinelanguage{lean}{
  morekeywords={theorem,lemma,def,structure,where,variable,section,namespace,
    end,open,in,if,then,else,fun,by,have,show,from,let,public,noncomputable,
    module,import,instance,class,abbrev,Prop,Type},
  sensitive=true,
  morecomment=[l]{--},
  morecomment=[n]{/-}{-/},
  morestring=[b]",
}
\lstnewenvironment{leandoc}{\lstset{
  language={},
  basicstyle=\small,
  keywordstyle={},
  commentstyle={},
  moredelim=[is][\ttfamily]{`}{`},
  literate={-}{{-}}1,
  frame=leftline,
  framerule=0.6pt,
  framesep=0.8em,
  xleftmargin=2.3em,
}}{}

\definecolor{topicbg}{HTML}{EEF3FA}
\definecolor{topicrule}{HTML}{3A6EA5}
  {\par\medskip\noindent\begingroup\color{topicrule}\textbf{\small #1}\endgroup
   \begin{itemize}[leftmargin=1.4em,itemsep=1pt,topsep=2pt,label=\textcolor{topicrule}{\textbullet}]\small}%
  {\end{itemize}\medskip}

\newcommand{\lean}{Lean~4}
\newcommand{\autoip}{\textsf{AutoIP}}
\newcommand{\mathlib}{\textsf{Mathlib}}
\newcommand{\tauceti}{\textsf{Tau Ceti}}

\newcommand{\Ll}{\mathcal{L}}
\newcommand{\Ran}{\mathcal{R}}
\newcommand{\Ker}{\mathcal{N}}
\newcommand{\Dom}{\mathcal{D}}
\newcommand{\dg}{^{\dagger}}

\title{Formalizing inverse problems in the age of AI:\\
       possibilities, pitfalls, and an invitation}
\author{Christian Clason\thanks{University of Graz, Austria}
\and Tapio Helin\thanks{LUT University, Finland}
   \and Medet Nursultanov\thanks{AI University, Kazakhstan}
   \and Lauri Oksanen\thanks{University of Helsinki, Finland}}

\date{Work in progress --- \today}

\begin{document}
\maketitle

% =====================================================================
\begin{abstract}
Proof assistants such as \lean{} check mathematical proofs down to the axioms, and AI agents
have recently made writing such proofs cheap. We describe \autoip{}, a \lean{} library in which
classical linear regularization theory, together with the functional analysis it rests on, has
been autoformalized by AI agents. A machine-checked proof certifies only the formal statement it
proves, so the remaining human task is to confirm that each formal statement says what was
intended. We organize the library around this task: a reviewer compares the formal statements
with their natural-language counterparts and never needs to read a proof. Our experience is that
the risks lie in the statements rather than in the proofs, and that simple automated tests on the
statements can expose problems even in the established literature. We illustrate this with a
textbook theorem whose hypothesis is more restrictive than asserted. We invite the inverse problems
community to extend and build on the library.
\end{abstract}


\tableofcontents   % \note{remove for submission}

% =====================================================================
\section{Introduction}
\label{sec:intro}

On 6 October 2026, OpenAI published more than seven hundred mathematical
manuscripts, all produced by an unreleased internal model~\cite{OpenAI2026}. A substantial number, though fewer
than half, of the released results come with a \emph{formal proof}
checked by Lean~\cite{deMoura2015, deMouraUllrich2021}.
Formal proof used to be very expensive. In 2008, Georges Gonthier wrote on the four color theorem that formalizing the infamously large case analysis in its proof was the easy part, done in a few
part-time months, while certain graph-theoretic arguments 
``took a few years to sort out''~\cite{Gonthier2008}. 
According to OpenAI, an average result took about
three hours of computation. The numbers cannot be compared directly. Not every result of OpenAI was
formalized, but the three hours also include finding the proof, which is usually the harder task.
Even with these caveats, the cost of a formal proof has fallen by orders of magnitude. 

Appel and Haken's original proof of the four color theorem in 1976
is often viewed as the first major
theorem whose proof no human could verify by hand. It is a reduction to almost two thousand configurations that were checked by a computer program~\cite{AppelHaken1977, AppelHakenKoch1977}. At the time, philosophers asked whether such an argument was a proof at all~\cite{Tymoczko1979}.
Accepting the original proof of the four color theorem meant trusting a custom program
written for that one purpose. Formal proof offers a way around this. The idea goes back to John McCarthy in
1962~\cite{McCarthy1962}: write the proof in a language precise enough that a single, small,
general-purpose program can check every step. A number of systems followed,
including Automath (1967)~\cite{deBruijn1968, deBruijn1970}, LCF (1972)~\cite{Milner1972}, and later Coq~\cite{CoquandHuet1988} and Lean, among others. 

The correctness of a formal proof does not depend on any
code written specifically for that proof. All trust rests on the proof checker, which is shared by all its
users and has been examined by them for years. In other words, trust is moved out of each
individual proof and into one piece of common infrastructure, and the correctness of a proof does not require a review.
What still requires review is the statement: the checker certifies that a proof establishes its
formal statement, not that the formal statement expresses the intended mathematics. This is the
main subject of the present paper.

The current technology-driven crisis in mathematics appears more profound than the controversy around Appel and Haken's proof in the 70s. AI systems now produce results at a volume that human refereeing cannot match, and the question of trust that the four color theorem raised for a single proof now arises for whole bodies of work. Nonetheless, we expect that formalization will become an increasingly important tool, since its cost has collapsed. Agents based on openly available frontier models can already formalize mathematics automatically with relative ease~\cite{Armstrong2026}. Such automatic formalization is often called \emph{autoformalization}~\cite{WangKaliszykUrban2018, Wu2022autoformalization}.  

While it is unknown according to which principles OpenAI decided what to formalize, a potential deciding factor could have been the size of the gap between already formalized theory and that required by a result. In the present paper we address this gap within the field of inverse problems. We describe \autoip{}, a Lean library of autoformalized results in linear regularization theory. Shared libraries not only speed up the formalization process but also save computational resources by eliminating overlapping work.
Inverse problems are a natural field for such a library. The classical theory rests on functional
analysis that is largely available in \mathlib{}, and it is now being extended rapidly by learned
and data-driven regularization methods. Convergence guarantees for such methods are new and less
scrutinized theory, where we would expect formalization to find more issues of the kind we report
in \cref{sec:nonvacuity} in the settled classical theory. 

Cheap formalization may change how applied mathematics is
done. An applied inverse problem is usually studied piece by piece: uniqueness and stability, regularization, discretization, convergence of the numerical method and
the model of the noise are treated in separate papers, often under assumptions that do not quite
fit together. A formal library can hold such pieces in one place and check that the conclusions
of one match the assumptions of the next. This makes large projects feasible that follow an
applied problem from its mathematical model to guarantees for the algorithm that solves it. Much
of the work such projects require, for example tracking constants, checking that every case is
covered, or verifying that a known result carries over to a slightly different setting, is tedious
and often simply not done. This is the kind of work that can now be
delegated to AI agents, while people decide what should be proved and check that it was.

\subsection{Reviewing formal libraries}

We hope that \autoip{} serves as an invitation for the inverse problems community to build and manage a formalized library of results specific to the field. While such results could in principle be incorporated in Lean libraries of general mathematics, there are benefits in keeping the review process within the community. The two existing general libraries, \mathlib{}~\cite{mathlib2020} and \tauceti{}~\cite{TauCeti}, sit at opposite ends of a spectrum regarding the review process. 

Most formalization in \lean{} builds on
\mathlib{}, a large library of formalized mathematics maintained by its
community under human review. It covers much of the standard curriculum, including the parts of
functional analysis that the theory of inverse problems rests on. Human reviewers of \mathlib{} not only attest the correctness of statements but also judge the quality of proofs. The review process is slow. 

\tauceti{} builds on \mathlib{} and adds formalized mathematics on top of it at great speed. The
speed comes from leaving the review to AI agents. They check each contribution, including whether
its statements and definitions say what they should, against roadmaps and review rubrics written
by humans.

We propose a middle ground where the correspondence of statements is attested by human reviewers, but proofs are not read.

\section{The \autoip{} library}
\label{sec:autoip}

\subsection{Relation to Mathlib and Tau Ceti}
\label{sec:relation}

Much of the functional-analytic material in Part~I of \autoip{} is available in
\mathlib{}. When \mathlib{} already contains a result, \autoip{} has no statement of its own but a citation to \mathlib{}. It is thereby also an index into the part of \mathlib{}
the theory rests on.  

The operator-theoretic infrastructure on top of
\mathlib{} has recently grown quickly in \tauceti{}. Its relevant human-written roadmap is \emph{Polar decomposition: functional calculus, the operator modulus, and singular
systems} \citep{TauCetiRoadmapPD}. Within a few days in October~2026, \tauceti{} merged several formalized results outlined in the roadmap, including the Moore--Penrose inverse \citep{TauCetiPR13805}. For maps
$T \colon X \to Y$ and $B \colon Y \to X$ between finite-dimensional inner product spaces,
$B$ is the Moore--Penrose inverse of $T$ if and only if it satisfies the four Penrose equations
\begin{equation}\label{eq:penrose}
  TBT = T, \qquad BTB = B, \qquad (TB)^* = TB, \qquad (BT)^* = BT.
\end{equation}

This material was largely imported from an
existing formalization developed at Kitware \citep{KitwareDKPS}. The Moore--Penrose
inverse and singular systems of \tauceti{} are, however, finite-dimensional objects,
with one exception: uniqueness under the Penrose equations \eqref{eq:penrose} is proved in arbitrary
inner product spaces. In particular, \tauceti{} represents a Moore--Penrose inverse as
an everywhere-defined linear map. This form only covers operators with closed range,
so it cannot express the ill-posed case. 

\autoip{} formalizes the
infinite-dimensional theory needed in inverse problems. It contains a singular
system $\{(\sigma_n, v_n, u_n)\}$ for every compact operator $K$ between Hilbert spaces,
with $\sigma_n$ nonincreasing and $\sigma_n \to 0$ when there are infinitely many,
and the expansion $Kx = \sum_n \sigma_n \langle v_n, x\rangle u_n$. It also contains
the Moore--Penrose inverse $T\dg$ of a bounded operator, given as a partially
defined linear map on $\Ran(T) \oplus \Ran(T)^\perp$ that sends $y$ to the
minimum-norm least-squares solution of $Tx = y$. Around it, \autoip{} includes the classical result that $T\dg$ is bounded exactly when $\Ran(T)$
is closed, the non-closed case being the one of interest to inverse problems. 
The definitions in \autoip{} and \tauceti{} can be connected, of course. In the closed-range
case, the operator $T\dg$ of \autoip{} satisfies the Penrose equations~\eqref{eq:penrose}, so by
uniqueness it coincides with \tauceti{}'s Moore--Penrose inverse wherever both are
defined.

\subsection{Scope}

\autoip{} builds the
regularization theory on top of functional analysis. At the time of writing, this theory has no overlap with \mathlib{} or \tauceti{} and \Cref{tab:size} summarizes the size of the library. The contents follow the structure of the linear part of the lecture notes~\citep{Clason2021}, which reorganize classical material from~\citep{EnglHankeNeubauer1996,Kirsch2011}. They cover the Moore--Penrose inverse, source conditions, parameter choice rules, convergence rates and order optimality, spectral, Tikhonov and Landweber regularization, and discretization as regularization. Four further chapters, drawing
more extensively on~\citep{EnglHankeNeubauer1996,MathePereverzev2003}, add general source conditions,
iterated Tikhonov and semi-iterative methods, perturbed operators, and the balancing
principle. The material is settled, and it rests on Hilbert space theory that is largely in
\mathlib{}, so the gap to existing formalized mathematics is small. 

There are many ways to extend the library. They naturally include nonlinear problems
as well as modern topics such as variational regularization in Banach spaces and statistical or machine-learning-based approaches to inverse problems. Some of these topics need foundations from other areas, such as convex analysis or statistics, that do not naturally belong in \autoip{} but are not yet available (to the required extent) in \mathlib{}.

  \begin{table}[t]
    \centering
    \caption{Size of \autoip{}.}
    \label{tab:size}
    \begin{tabular}{lr}
      \toprule
      Definitions & 51 \\
      \quad cited from \mathlib{} & 14 \\
      \quad predicates & 15 \\
      \quad constructions & 22 \\
      \quad\quad reviewed by an author & 22 \\
      Theorems & 73 \\
      \quad cited from \mathlib{} & 22 \\
      \quad stated in Lean & 51 \\
      \quad\quad reviewed by an author & 25 \\
      Lemmas / corollaries / propositions & 83 / 16 / 5 \\
      Lean files (library / tests) & 523 / 152 \\
      Lines of Lean (library / tests) & 22\,312 / 13\,250 \\
      Non-vacuity obligations & 503 \\
      \bottomrule
    \end{tabular}
  \end{table}

\section{Organization of the library}
\label{sec:design}

Lean checks that proofs establish the formal statements
they are attached to. It does not attest that the statements say what was
intended. Our review model asks a person to read the formal statement
against the informal one, and the architecture of \autoip{} is organized around
it: a minimal, fixed set of files is designated for reading, and everything else
is arranged so that it need not be read unless the reviewer is interested in the particulars of the proofs. 
A formalization aimed at one main result has one statement to compare, together
with the definitions it uses. In the formalization of Fermat's Last Theorem
\citep{AnthropicFLT2026}, for example, the final statement is compared with
the one stated in \mathlib{}, which unfolds to four lines of definitions. A library has no main result. Each definition and
each theorem is a deliverable, and each has to be compared with its informal
counterpart. 

The natural language statements are given in an ordinary \LaTeX{} document, which we call the blueprint and whose definitions, lemmas, and theorems we call nodes. Each node corresponds to one Lean file, its
\emph{specification}, at a path determined by the position of the node in the
blueprint. The specification contains the formal statement of the node and
no other statements. A reader attesting the correctness of the formalized statements works from the PDF of the blueprint and goes from a node
to its specification. The remaining Lean files of the library are not part of
what they read. A companion extension for Visual Studio Code~\cite{Sini} supports this
workflow. A click on the label of a node in the blueprint source opens its specification, and a
key binding in a Lean file shows the corresponding statement in the PDF. Together with the Lean
extension, which follows the citations of \mathlib{} in the specifications, the reader can move
between the blueprint, the formal statements and \mathlib{} without searching.

\Cref{fig:node} shows a node and its specification. Let $K \in \Ll(X,Y)$ be compact, with
singular system $\{(\sigma_n, v_n, u_n)\}_{n \in I}$, and let $\kappa := \|K\|^2$. A function
$\varphi$ that is bounded on $[0,\kappa]$ defines a bounded operator on $X$ by
\[
  \varphi(K^*K)x := \sum_{n \in I} \varphi(\sigma_n^2) \langle v_n, x\rangle v_n
    + \varphi(0) P_{\Ker(K)} x .
\]
Here $P_{\Ker(K)}$ is the orthogonal projection onto the kernel of $K$.
A family of such functions $\{\varphi_\alpha\}_{\alpha > 0}$, called a filter, defines the
spectral regularization $R_\alpha := \varphi_\alpha(K^*K)K^*$, so that
$R_\alpha y = \sum_n \varphi_\alpha(\sigma_n^2)\, \sigma_n \langle u_n, y\rangle v_n$. The Tikhonov
filter is $\varphi_\alpha(\lambda) := (\lambda + \alpha)^{-1}$. The node of
\cref{fig:node} states that $R_\alpha y$ is the unique solution of the regularized normal equation.

\begin{figure}[t]
  \begin{quote}
    \textbf{Lemma} (Tikhonov solves a regularised normal equation). Let $y \in Y$ and
    $\alpha > 0$. For $x \in X$ one has $x = R_{\alpha}y$ if and only if
    $(K^{*}K + \alpha\,\mathrm{Id})x = K^{*}y$.
  \end{quote}
\begin{lstlisting}
theorem tikhonov_iff_normal (hs : IsSingularSystem K σ v u) {α : Set.Ioi (0 : ℝ)}
    (y : F) (x : E) :
    x = spectralReg hs tikhonovFilter (tikhonovFilter_exists_bound_on_spectrum hs) α y ↔
      (K.adjoint ∘L K + ((α : ℝ) : 𝕜) • ContinuousLinearMap.id 𝕜 E) x = K.adjoint y
\end{lstlisting}
  \caption[A node of the blueprint and its specification]{A node of the blueprint (top) and
    the formal statement in its specification (bottom). The proof is omitted. In the Lean code
    listings of this paper, \texttt{$\Bbbk$} is the scalar field, $\mathbb{R}$ or $\mathbb{C}$,
    and \texttt{E} and \texttt{F} are the Hilbert spaces $X$ and $Y$. Here \texttt{hs} is the
    singular system of $K$, \texttt{$\alpha$ : Set.Ioi (0 : $\mathbb{R}$)} is a parameter
    $\alpha > 0$, and \texttt{spectralReg hs tikhonovFilter \_ $\alpha$ y} is $R_\alpha y$. Its
    third argument is a proof that the Tikhonov filter is bounded on $[0,\kappa]$, which the
    functional calculus requires. On the right-hand side of the equivalence, \texttt{K.adjoint $\circ$L K} is
    $K^*K$ and \texttt{ContinuousLinearMap.id $\Bbbk$ E} is the identity on $X$.}
  \label{fig:node}
\end{figure}

The choice of organization was inspired by the formalization of Fermat's Last Theorem. This was carried out on Prove2Me
\citep{Prove2Me2026}, which uses separate files for each statement. There the separation serves compilation speed
and the coordination of many agents, and trust rests on the final statement.
In \autoip{} it also serves the reader. A similar architecture is used by the recent formalization harness Trellis~\cite{Trellis}. 

A critical part of a formalization effort is the choice of definitions. In Lean, definitions are often constructive, and mathematically equivalent
constructions can differ greatly in how usable they are. For this reason, \autoip{} refers to concrete constructions only locally, and uses them globally via characterizing properties. This is enforced by the module
system of Lean, which separates what a file exports from what it uses internally. A file that
imports a definition sees its name and type but, unless the definition is explicitly exposed, not
its body. In \autoip{} only one file per node, its \emph{API}\footnote{Shorthand for application programming interface.} holding the characterizing properties, imports the
specification with access to the bodies. Every other node sees the definition only through those
properties.

\section{How the work was carried out}
\label{sec:process}

The blueprint was written by the authors, its statements following~\citep{Clason2021,
EnglHankeNeubauer1996}, and the Lean code was written by AI agents. The first round of
autoformalization was carried out with few instructions beyond the blueprint itself. Reviewing
its outcome afterwards led to several rounds of restructuring, some of which are described in the
next section. Our working method grew out of those rounds and was not in place
during the first one. We compare it with the workflow of Armstrong~\citep{Armstrong2026} in
\cref{sec:armstrong}. As far as possible, our practices are implemented as programmatic checks. For example, the files of the library must follow the organization outlined in Section~\ref{sec:design}, no axioms can be introduced, and no
unused code is allowed to remain. An agent works until these checks pass. 

What no program can check is recorded instead. For each node, a record states whether its
specification says what the node says, and whether this was read by an author or by an agent. For
each constructive definition, the agent records why it chose its construction over
the alternatives, and an author approves or rejects the choice. A rejection or a mismatch stands
until the text it objects to changes. Each record carries hashes of the informal and the formal
statement it was written about, and of everything these rest on: the nodes the prose refers to and
the definitions the formal statement uses. 
Formal statements use a hashing scheme similar to that of
Trellis~\cite{Trellis}. 
When either hash changes, the record expires and the
node is marked for a new reading, while a change of a proof leaves the record intact. This made the
repeated restructuring affordable, since the records showed exactly which statements had to be
read again.

\section{Lessons learned}

\subsection{Delimiting nodes is helpful}
\label{sec:delimiting}

Our first attempt at organizing the results in \autoip{} took the shape of a conventional
library, inspired by \mathlib{}, with formalizations of distinct nodes sharing a file. Thus the set of statements that required comparison was not delimited. 
Such organization made attestation of the correspondence between informal and formal statements cumbersome for us. Moreover, reviews by successive AI agents disagreed.
For instance, GPT-5.6 Sol reviewed the library against slightly adapted\footnote{\tauceti{} uses the rubrics for agentic review of single pull requests. We adapted them to library-wide review.} code-quality (generality, API design, naming, placement, proof quality, documentation and reuse)
rubrics of \tauceti{}~\cite{TauCetiReview}
and raised about 300 objections. Claude Opus~5, the model that wrote the code,
acted on slightly over half of them and rejected the rest of the findings. When Claude
Opus~5 then reviewed the material against the same rubrics, it raised more than 450 new objections. We stress that the model objected to the already revised code written by the same model in a separate session.

The reviews of correspondence between formal and informal statements were more stable but not
settled: Claude Opus~5 judged 17 of the 224 nodes of the library at that time a mismatch. After the
restructuring into the organization described above, three rounds of such reviews by Claude Opus~5.5 covered between 205 and 229 nodes each and found a single mismatch in total. (The number of nodes varies because the library changed between the rounds and one round did not cover all of it.) We attribute the improvement in part to the restructuring, which gave each review a delimited task, although the change of model may also have contributed.

\subsection{Junk values are dangerous}
\label{sec:junk-values}

Every function in Lean is total. A function that is meaningful only on part of its
natural domain must still return a value elsewhere, and \mathlib{} returns an arbitrary value, often called a
\emph{junk value}. For example, \texttt{Module.finrank}, the dimension of a vector space as a
natural number, is $0$ on every infinite-dimensional space. Junk values are convenient when writing proofs by hand, since proper treatment of the out-of-domain values would lead to longer arguments. But they come with a cost: it is easy to write a formal statement that says less than intended. Our simplest example is the following lemma of the blueprint.

\begin{quote}
  \textbf{Lemma} (Dimension counting). Let $X$ be a Hilbert space and let $U, V \subseteq X$
  be subspaces with $U$ finite-dimensional and $\dim U < \dim V$. Then
  $V \cap U^\perp \ne \{0\}$.
\end{quote}
It was initially formalized as
\begin{lstlisting}
theorem Submodule.exists_ne_zero_mem_inf_orthogonal
    {U V : Submodule 𝕜 E} [U.HasOrthogonalProjection] [FiniteDimensional 𝕜 U]
    (h : Module.finrank 𝕜 U < Module.finrank 𝕜 V) :
    ∃ v : E, v ≠ 0 ∧ v ∈ V ∧ v ∈ Uᗮ
\end{lstlisting}
If $V$ is infinite-dimensional, then \texttt{Module.finrank} $\Bbbk$ \texttt{V} is $0$ and the hypothesis
\texttt{h} fails. The formal lemma therefore says nothing about infinite-dimensional $V$,
although the informal one allows it. 
The current formalization compares dimensions as cardinals, which have no junk
value on infinite-dimensional spaces:\footnote{The hypothesis \lstinline|[U.HasOrthogonalProjection]| of the first version, which asks for the
orthogonal projection onto $U$ to exist, is absent from the second only because it follows from
\lstinline|[FiniteDimensional 𝕜 U]| and Lean infers it automatically.
}

\begin{lstlisting}
theorem Submodule.exists_ne_zero_mem_inf_orthogonal_of_rank_lt
    {U V : Submodule 𝕜 E} [FiniteDimensional 𝕜 U]
    (h : Module.rank 𝕜 U < Module.rank 𝕜 V) :
    ∃ v : E, v ≠ 0 ∧ v ∈ V ∧ v ∈ Uᗮ
\end{lstlisting}
The old statement is not vacuous: its hypotheses can be satisfied, and the non-vacuity tests
described below would not have caught it. It is simply weaker than the informal lemma. To read
a formal statement correctly, a reviewer must know the junk value of every function in it. \autoip{} aims to introduce no new junk values by defining each function on its natural domain. Its statements also carry hypotheses that rule out \mathlib{}'s junk values. The resulting longer proofs are of little consequence, since the proofs are autoformalized. 

\mathlib{} documents its junk values. In the case of \texttt{Module.finrank} it says

\begin{leandoc}
The rank of a module as a natural number.

For a finite-dimensional vector space `V` over a field `k`, `Module.finrank k V` is equal to
the dimension of `V` over `k`.

For a general module `M` over a ring `R`, `Module.finrank R M` is defined to be the supremum of the
cardinalities of the `R`-linearly independent subsets of `M`, if this supremum is finite. It is
defined by convention to be `0` if this supremum is infinite. See `Module.rank` for a
cardinal-valued version where infinite rank modules have rank an infinite cardinal.
\end{leandoc}

\subsection{Bad prose is confusing}
\label{sec:bad-prose}

An informal definition written for formalization should
state the type of the object it defines and should live in a single definition environment in the blueprint.
In the first version of the blueprint, the Moore--Penrose inverse violated these principles: it had no definition of its
own; rather, it was introduced in a theorem.

\begin{quote}
  \textbf{Theorem} (Existence and uniqueness of the Moore--Penrose inverse). Let
  $T \in \Ll(X,Y)$ and let $\tilde{T} := T|_{\Ker(T)^{\perp}} \colon \Ker(T)^{\perp}
  \to \Ran(T)$. Then $\tilde{T}$ is bijective, and $\tilde{T}^{-1}$ has a unique linear
  extension $T\dg$ with
  \[ \Dom(T\dg) = \Ran(T) \oplus \Ran(T)^{\perp}, \qquad \Ker(T\dg) = \Ran(T)^{\perp}. \]
  $T\dg$ is called the \emph{Moore--Penrose inverse} of $T$. It is in general unbounded.
\end{quote}
A second characterization, as a minimum-norm solution, was given in a subsequent theorem.

\begin{quote}
  \textbf{Theorem} (The Moore--Penrose inverse computes the minimum norm solution). Let
  $y \in \Dom(T\dg)$. Then
  (i)~least squares solutions of $Tx = y$ exist, and are exactly the solutions of
  $Tx = P_R y$;
  (ii)~there is a unique minimum norm solution, namely $x\dg := T\dg y$;
  (iii)~the set of all least squares solutions is $x\dg + \Ker(T)$.
\end{quote}
Here $P_R$ is the orthogonal projection onto $\overline{\Ran(T)}$. The autoformalization took the second characterization, which does not say explicitly that $T\dg$ is a linear map. The resulting formal definition did not make $T\dg$ a linear map on its domain.
\begin{lstlisting}
def mpDomain (T : E →L[𝕜] F) : Submodule 𝕜 F :=
  LinearMap.range T.toLinearMap ⊔ (LinearMap.range T.toLinearMap)ᗮ

noncomputable def moorePenrose (T : E →L[𝕜] F) (y : F) : E :=
  if h : ∃ x, IsMinimumNorm T y x then h.choose else 0
\end{lstlisting}
Here \lstinline|E →L[𝕜] F| is the space of bounded linear operators from \lstinline|E| to
\lstinline|F|, and \texttt{moorePenrose} is a function on all of \texttt{F}, with junk value $0$ outside the domain. Its
linearity is two separate lemmas, each carrying the hypothesis $y \in \Dom(T\dg)$. Every
use of $T\dg$ had to carry the same hypothesis and re-derive the same facts. 

The current blueprint defines the inverse in one place.

\begin{quote}
  \textbf{Definition} (The Moore--Penrose inverse). Let $T \in \Ll(X,Y)$. The
  \emph{Moore--Penrose domain} of $T$ is
  $\Dom(T\dg) := \Ran(T) \oplus \Ran(T)^{\perp} \subseteq Y$, and the \emph{Moore--Penrose
  inverse} $T\dg$ of $T$ is the linear map with domain $\Dom(T\dg)$ sending $y$ to the
  minimum norm solution of $Tx = y$ in the sense of the definition of least squares and minimum norm solutions. Outside that domain
  $T\dg$ takes no value. It is in general unbounded.
\end{quote}

\noindent The formal definition follows it, as a linear map on $\Dom(T\dg)$, packaged as a
partially defined linear map from $Y$ to $X$ (the construction, still based on the minimum norm solution, and proofs of linearity are omitted).
\begin{lstlisting}
noncomputable def moorePenroseLinear [CompleteSpace E] (T : E →L[𝕜] F) :
    mpDomain T →ₗ[𝕜] E where ...

noncomputable def moorePenrose [CompleteSpace E] (T : E →L[𝕜] F) : F →ₗ.[𝕜] E :=
  ⟨mpDomain T, moorePenroseLinear T⟩
\end{lstlisting}
Here \lstinline|mpDomain T →ₗ[𝕜] E| is the space of linear maps from the domain to
\lstinline|E|, and \lstinline|F →ₗ.[𝕜] E| that of partially defined linear maps from
\lstinline|F| to \lstinline|E|.

\subsection{Non-vacuity tests catch mistakes}
\label{sec:nonvacuity}

A theorem whose hypotheses are contradictory is provable and states nothing. A
theorem whose hypotheses always hold may be read as stating more than it does.
To guarantee that \autoip{} does not contain such vacuous statements, for every statement in a specification, two obligations were generated using a deterministic (not AI-based) algorithm: exhibit an instance in which the hypotheses hold, and one in which they fail. AI agents were then instructed to prove that the obligations hold. 

The satisfiability obligations exposed a theorem from the canon of the regularization theory literature whose hypothesis is far more restrictive than it appears. 
Let $T \in \Ll(X,Y)$ be a compact operator between Hilbert spaces, and for $\nu, \rho > 0$
consider the source sets
\[
  X_{\nu,\rho} := \{x \in X : \exists w \in X,\ x = (T^*T)^{\nu/2} w,\ \|w\| \le \rho\}.
\]
Let $\{R_\alpha\}_{\alpha > 0}$ be a family of maps $Y \to X$, for example
$R_\alpha = (T^*T + \alpha I)^{-1} T^*$ in Tikhonov regularization. A \emph{parameter choice}
is a map $\alpha \colon (0,\infty) \times Y \to (0,\infty)$ that selects the parameter
$\alpha(\delta, y^\delta)$ from the noise level $\delta$ and the data $y^\delta$. It is
\emph{a priori} if it depends on $\delta$ alone. The pair $(R_\alpha, \alpha)$ is a
\emph{convergent regularization method} if $E(y,\delta) \to 0$ as $\delta \to 0$ for every
$y \in \Dom(T\dg)$, where
\[
  E(y,\delta) := \sup_{\|y^\delta - y\| \le \delta}
    \bigl\| R_{\alpha(\delta, y^\delta)} y^\delta - T\dg y \bigr\|
\]
is the worst-case error. The pair is \emph{order optimal} for $\nu$ and $\rho$ if there is
$c \ge 1$ such that
\[
  E(Tx, \delta) \le c\, \delta^{\nu/(\nu+1)} \rho^{1/(\nu+1)}
  \qquad \text{for all } \delta > 0 \text{ and } x \in X_{\nu,\rho},
\]
see \citep[Def.~3.17]{EnglHankeNeubauer1996} and \citep[(4.8)]{Clason2021}. Parameter choices
such as the discrepancy principle compare the residual with $\tau\delta$ for a fixed
$\tau > 1$, and the theorem below is stated for the whole family of rescaled choices
$\alpha_\tau(\delta, y^\delta) := \alpha(\tau\delta, y^\delta)$. Both sources state that if,
for every $\tau > \tau_0$, the pair $(R_\alpha, \alpha_\tau)$ is order optimal for a fixed
$\nu$ and all $\rho > 0$, then it is a convergent regularization method
\citep[Thm.~3.18]{EnglHankeNeubauer1996}, \citep[Thm.~4.14]{Clason2021}. This abstract convergence theorem is true,
but its hypothesis is far stronger than it looks. 

Take $x = 0$, which lies in $X_{\nu,\rho}$
for every $\rho$, and data $y^\delta$ with $\|y^\delta\| \le \delta$. The reconstruction
$R_{\alpha_\tau(\delta, y^\delta)} y^\delta$ is then at distance at most
$c\, \delta^{\nu/(\nu+1)} \rho^{1/(\nu+1)}$ from $x = 0$ for every $\rho > 0$. Since $c$ does
not depend on $\rho$, letting $\rho \to 0$ shows that the method returns $0$ on all data with
$\|y^\delta\| \le \delta$. The hypothesis asks for one parameter choice that works for all
$\rho$, so a priori rules tuned to a given $\rho$ are not admitted. For an a priori parameter choice, where $\alpha_\tau$ depends on
$\delta$ alone and $R_\alpha$ is linear, this means $R_{\alpha_\tau(\delta)} = 0$ for every
$\delta$, so the method can converge only if $T\dg = 0$, that is, $T = 0$. Likewise, if the
parameter choice takes values in $(0,\infty)$, as in our definition, a family of injective
$R_\alpha$, such as Tikhonov regularization, returns $0$ only on $y^\delta = 0$, so it satisfies
the hypothesis only if $Y = \{0\}$.

The discrepancy principle of \citep{EnglHankeNeubauer1996} avoids the second restriction: on data
with $\|y^\delta\| \le \tau\delta$ it chooses $\alpha = \infty$ and returns $0$, which is exactly
what the hypothesis demands. The application of the abstract convergence theorem to the discrepancy principle there
\citep[Thm.~4.17]{EnglHankeNeubauer1996} is therefore sound. A priori rules, however, cannot
satisfy the hypothesis, although right after the proof of the theorem
\citep[p.~63]{EnglHankeNeubauer1996} states that ``the assumptions of Theorem~3.18 are fulfilled
for all parameter choice strategies that will be considered in this book''. The lecture
notes~\citep{Clason2021} define the discrepancy principle without the value $\alpha = \infty$,
and use the theorem to prove convergence both for a priori rules and for the discrepancy
principle \citep[Thms.~5.9 and~5.11]{Clason2021}. These appeals to the theorem fail, although the
convergence results themselves remain true.

The abstract convergence theorem was initially formalized and proved in \autoip{}, with parameter choices taking values
in $(0,\infty)$ as in~\citep{Clason2021}. Its satisfiability obligation was first
discharged with $T = 0$ on the trivial space, which raised no alarm. However, when the non-vacuity witnesses were later required to be non-trivial, this obligation was the only one that kept
the zero operator. An agent review then explained why the hypothesis can hold only in degenerate cases, and the theorem was removed from the library.

No formal proof in \autoip{} had ever used the theorem, while the agents were given natural language proofs, following~\citep{Clason2021}, that deduce the
convergence of, for example, Tikhonov regularization from it. The autoformalization produced correct convergence proofs that bypassed the theorem. Aligned statements guarantee that the theorems proved are the intended ones. They say nothing
about whether a formal proof follows an informal one. In this case the divergence hid a gap in
the informal proofs: the appeals to the abstract convergence theorem fail, although the theorems they serve remain true. That a verified formalization need not validate the informal proof it started from is
discussed further in~\cite{Bastounis2026}.

\section{Comparison with Armstrong's workflow}
\label{sec:armstrong}

A detailed account of an autoformalization workflow is given by
Armstrong~\citep{Armstrong2026}, together with the instructions he gives to his agents~\citep{ArmstrongSkills}.
It is probably the most thorough account available at the time of writing. These instructions were published in
October~2026, when our work had largely been carried out, which is why we did not follow them more
closely. Nonetheless, our approach shares the basic principles with that of Armstrong. Proofs
may be left unfinished while statements are settled, but definitions may not, and a definition must not introduce junk values.
A person signs off on the correspondence between a formal statement and its informal counterpart, and such attestations are recorded. Beyond this, the two
approaches differ in ways that reflect the material they were applied to.

Armstrong formalizes research papers, where the route of a proof may need
to be chosen, steps are left implicit, and the source may contain errors. His agents first extract
from the paper a dependency graph that records the results, the substantive steps of their proofs,
alternative routes, and the places where the argument has gaps. \autoip{} consists of textbook material
with clear dependencies. The blueprint already gives such a graph, its nodes connected by their
cross-references. The graph can be extracted programmatically using a \LaTeX{} parser such as plas\TeX~\cite{plasTeX}, and a separate graph would be redundant for us.

Formalizing a research paper often leads to a deep tree of statements with the root corresponding
to the main result of the paper, while \autoip{} is a forest of statements that form shallower
chains. Armstrong selects a small set of key statements that are approved by the author before any
proof work starts and then frozen. A statement that turns out to be wrong is replaced by a new
version, with renewed approval. Freezing guards against a statement being quietly weakened to make its
proof go through. Consquences of such a weakening are smaller in a shallow forest than in a deep tree, since fewer statements are affected by it. We deferred reviewing
statements until after they had been proved. On the one hand, no reading was spent on statements
that were later changed or discarded while the proofs were written. On the other hand, bad design
choices by an agent went unnoticed until they had global consequences. For instance, repairing the
awkward formalization of the Moore--Penrose inverse, described in \cref{sec:bad-prose}, required
several hours of agent compute.

Armstrong's agents audit each statement by a protocol that establishes the correspondence between premises of the formal statement and those of the informal
one. Each hypothesis and
argument must be justified as a piece of the informal text, part of
setting up the types, or a correction approved by the author.
Such a procedure did not occur to us. In turn, we generate non-vacuity tests
programmatically, and Armstrong's workflow seems to have no systematic
counterpart to them. The two kinds of checks complement each other. The abstract convergence
theorem of \cref{sec:nonvacuity} would pass Armstrong's classification, since its hypothesis is
taken from the informal source. It was caught by a non-vacuity test. Conversely, our non-vacuity
tests cannot detect a hypothesis that is satisfiable but should not be there. Neither kind of test
would catch the dimension counting lemma of \cref{sec:junk-values}, whose hypothesis corresponds to
one in the source text but is distorted by the junk value of a library function.

\section{An invitation to contribute}
\label{sec:invitation}

The review model of \autoip{} places the human effort where human judgement is needed: in
deciding whether a formal statement captures the meaning of the informal one. This is also its present
bottleneck. Of the roughly two hundred nodes of the library, all have been attested by AI
agents, but only some by the authors so far. 
The human-reviewed nodes include all definitions that construct new objects (the remaining definitions either cite \mathlib{} or define predicates). Reading a specification against its node in
the blueprint requires familiarity with the mathematics and the ability to read Lean statements,
but no experience in writing Lean proofs. We therefore see such attestation as the most valuable contribution, and also as the natural way into the library.

A second way to contribute is to extend the blueprint. A new chapter, written as ordinary
\LaTeX{} in the style of the existing ones, can be formalized by AI agents along the lines
described above. Natural candidates are results that are classical but absent from the present
selection, such as those related to regularization of nonlinear problems or to Bayesian inverse
problems. Some of
these need foundations that are not yet formalized, and part of the work is to identify them.

Finally, some of the infrastructure in \autoip{} is of general interest and belongs in general
libraries. The singular systems of compact operators and the Moore--Penrose inverse with its
natural domain, discussed in \cref{sec:relation}, are examples. Moving such material into
\mathlib{} or \tauceti{}, and replacing it in \autoip{} with citations, keeps the library focused
on the theory specific to inverse problems.

Contributions of this kind require shared conventions on review, credit and the disclosure of AI
use. We do not try to settle them here. Instead, we invite the community to discuss them. A possible forum is the \autoip{} repository,
\url{https://github.com/l-oksanen/autoip/discussions}.

\section{Discussion and outlook}
\label{sec:discussion}

\autoip{} covers classical material that has been settled for decades, and its blueprint
follows mainly two sources~\citep{Clason2021,EnglHankeNeubauer1996}. A library of this kind
does not show that research mathematics in inverse problems can be formalized as easily.
Nonetheless, even this material contained a theorem whose hypothesis is far more restrictive than
stated, and which is applied where the hypothesis does not hold. It was found not by reading but by a mechanical test, and
then only once the test was made strict enough. We expect that formalizing more recent and less
scrutinized work will expose more such issues, and we see this as one of the main benefits of
formalization for the field.

The lessons above also show what formalization does not provide. A machine-checked proof
certifies its formal statement and nothing else. Statements can be weakened by junk values,
definitions can be formalized in a shape that is technically correct but awkward to use, and a
formal proof can diverge from the informal proof it was meant to follow without anyone noticing.
None of these issues is detected by the proof checker, and all of them were found by people or
by tests designed by people, with agents helping to diagnose them. The human work therefore shifts from checking proofs to specifying
statements and reviewing them. This work requires deciding what was meant, and it is not clear
that it becomes cheaper as the models improve. The
disagreements between AI reviewers reported above, and the fact that an agent review accepted the
first formalization of the Moore--Penrose inverse, suggest that it cannot simply be delegated to
them either.

The cost of formalization will likely continue to fall. If it does, we expect formal proofs to
become a routine companion of research papers in inverse problems, as numerical code is today,
and formal libraries to become part of the infrastructure of the field. For this to happen, the
community needs shared libraries, shared conventions for what a reviewer must check, and people
willing to do the review.

% =====================================================================
\section*{Data availability statement}
The \lean{} source code and \LaTeX{} blueprint are openly available at
\url{https://github.com/l-oksanen/autoip}. 
%\note{archive a tagged release (e.g.\ Zenodo DOI) and cite the exact commit}.

\section*{Use of AI tools}
The \lean{} code of \autoip{} was written by AI agents: Claude Opus~5 and, from late September
2026, Claude Opus~5.5 (Anthropic). The same agents processed the proofs in the blueprint, whose
statements follow~\citep{Clason2021,EnglHankeNeubauer1996}. The code was reviewed by AI agents, GPT-5.6 Sol
(OpenAI) and Claude Opus~5 and~5.5, as described in the section on lessons learned. The
non-vacuity obligations were generated by a deterministic program and proved by Claude Opus~5.5. All
proofs are checked by the \lean{} kernel, and the correspondence between formal and informal
statements is attested as described in \cref{sec:design}.
The development was carried out with monthly subscriptions to the Claude Team plan for
scientists. A single premium seat for three months would have sufficed for the entire
development, although in practice most authors used plans of their own. The reviews by GPT-5.6
Sol used a single monthly ChatGPT Plus subscription.

In preparing this manuscript, Claude Opus~5.5 was used to search the version history of the
repositories and the \tauceti{} project for the data reported here, to draft parts of the text and
the bibliography, and to typeset the Lean listings. The authors reviewed and edited all text and
take full responsibility for the content.

\section*{Acknowledgements}

L.O. was supported by the European Research Council 
of the European Union, grant 101086697 (LoCal), 
and the Research Council of Finland grant 359182 (Flagship of Advanced Mathematics for Sensing Imaging and Modelling). 
Views and opinions expressed are those of the authors only and do not 
necessarily reflect those of the European Union or the other funding 
organizations.

% =====================================================================
\bibliographystyle{abbrvurl}
\bibliography{autoip_paper}

\end{document}